\pdfoutput=1

\documentclass[11pt]{article}

\usepackage{ACL2023}

\usepackage{times}
\usepackage{latexsym}

\usepackage[T1]{fontenc}

\usepackage[utf8]{inputenc}

\usepackage{microtype}

\usepackage{inconsolata}

\usepackage{cleveref}
\usepackage{graphicx}
\usepackage{booktabs}
\usepackage{subfigure}
\usepackage{multirow}
\usepackage{makecell}
\usepackage{enumitem}

\usepackage{xcolor}
\usepackage{colortbl}
\definecolor{backgroundgray}{gray}{.9}

\newcommand{\paragraphnew}[1]{\vspace{0.1cm} \noindent \textbf{#1}}

\title{Which Model Shall I Choose? \\Cost/Quality Trade-offs for Text Classification Tasks}

\author{Shi Zong$^1$\quad Josh Seltzer$^2$\quad Jiahua (Fiona) Pan$^2$\quad Kathy Cheng$^2$\quad Jimmy Lin$^1$\\
$^1$University of Waterloo \quad
$^2$Nexxt Intelligence \\
{\tt \{s4zong, jimmylin\}@uwaterloo.ca}\\
{\tt \{josh, fiona, kathy\}@nexxt.in}
}

\begin{document}

\maketitle

\begin{abstract}
Industry practitioners always face the problem of choosing the appropriate model for deployment under different considerations, such as to maximize a metric that is crucial for production, or to reduce the total cost given financial concerns.
In this work, we focus on the text classification task and present a quantitative analysis for this challenge.
Using classification accuracy as the main metric, we evaluate the classifiers' performances for a variety of models, including large language models, along with their associated costs, including the annotation cost, training (fine-tuning) cost, and inference cost.
We then discuss the model choices for situations like having a large number of samples needed for inference.
We hope our work will help people better understand the cost/quality trade-offs for the text classification task. 
\end{abstract}

\section{Introduction}

There have been great advances in building real-world natural language processing applications.
The success is mostly built on deep neural networks, which typically require a large amount of annotated supervised data for training \cite{10.1145/3439726}. 
Built on these neural architectures, researchers have taken a step further in training large language models \cite{radford2019language, JMLR:v21:20-074}.
These large language models have demonstrated great performance under few-shot or even zero-shot settings, where only a small amount of annotated data is needed \cite{NEURIPS2020_1457c0d6}.
It has emerged as a promising approach for many downstream applications.

Given this new trend, a natural question to ask is: after seeing that large language models, such as GPT-3, work at least reasonably well given only a handful of examples, whether one should directly use it in practice or still need to train one's own model? 
Such questions are particularly crucial for industry practitioners, as they often need to consider a number of factors when deploying a model.
For example, industry practitioners need to reach a balance between reasonable model performances and a relatively low cost.
Specifically, a clear picture that covers what kind of model along with how much annotation is needed to get a high performance, and what the cost implications are, is beneficial for making an informed model selection.

In this paper, we provide a thorough empirical study of the trade-offs between costs and quality for text classification models.
Our choice of the text classification task is inspired by our Inquisitive Natural Conversation Agent (INCA) platform \cite{10.1145/3477495.3536326}, which provides real-world value for identifying users' information intent during their interactions with a virtual agent.
We experiment with a representative selection of models, ranging from non-neural models (logistic regression and support vector machine) and traditional neural models (convolutional neural network) to the latest pre-trained language models (BERT, T5, and GPT-3). 
We evaluate the classification accuracy of these classifiers under a few-shot learning setting, so as to simulate situations of having only limited resources to annotate a certain number of samples. 
We then analyze the choices for the model selection by examining the cost/quality trade-offs in several different cases, such as having a large number of test samples. 

In sum, our paper studies the empirical question that is shared by many industry practitioners: \textit{Is it possible to achieve the same accuracy with a cheaper per-inference model, and if so, roughly how many annotation samples are needed?}
We hope our work will be a useful guide for practitioners to make informed decisions in selecting text classification models.

\section{Factors in Cost/Quality Trade-offs}
\label{sec:factors}

We start by providing a general discussion of the factors related to model costs and performance measurements, along with methods to estimate them.
We then consider a specific text classification task in \Cref{sec:setup} and present a detailed calculation for these factors in \Cref{sec:analysis}.

\subsection{Annotation Cost}

We use \textit{annotation cost} to describe the human efforts of assigning labels to data samples.\footnote{All costs in currency in this work are in US Dollars.}
As models normally rely on a certain amount of supervised training data, it is necessary to determine the number of annotated examples needed for production in the industry use case.

The annotation cost is affected by many factors, including the difficulty level of the annotation task, the type of annotation performed, and the person who will actually perform the annotation.
Specific to our classification task, simple tasks such as binary sentiment detection normally require less human effort, as they are straightforward and the labels are less ambiguous. 
Crowdsourcing or online labeling services are both suitable for these tasks. 
While for tasks that require in-domain or background knowledge, for example, our INCA Information Intent dataset (to be discussed in \Cref{sec:datasets}), in-house experts are normally required for annotation (or as a second step quality check). 
This annotation process would take a longer time and will lead to a higher total cost.

\subsection{Model Cost}

The \textit{model cost} is broadly defined as the cost for getting a trained model and making the model predict labels to examples.
In this work, we limit the concept of model cost to the following training/pre-training cost and inference (or evaluation) cost.

\paragraphnew{Training Cost.}
The training cost has different content for different types of models. 
For non-neural methods or traditional neural methods (i.e., no pre-training involved), the training cost only refers to the cost of training these classifiers.
For the pre-trained models, the training cost includes the pre-training cost and fine-tuning cost.\footnote{In this work, we do not consider the models' pre-training cost (discussed in \Cref{sec:cost_estimate}).}

\paragraphnew{Inference Cost.}
The evaluation cost estimates the cost for a trained model to predict labels for a given example. 
Factors such as model size, inference latency, and energy usage all contribute to the inference cost.

\paragraphnew{Cost Estimation.}
The model cost can be estimated based on the time length of the training and inference process, such as TPU or GPU hours, and some auxiliary costs such as data storage \cite{10.1145/3381831}.
For models that have already been pre-trained and do not report the actual running time, the total training cost can be estimated by using established training protocols, including the model size (number of parameters), the learning rate, number of epochs for parameter updates.

In recent years, companies like OpenAI, Hugging Face, and Cohere provide access to large language models through API calls to do inference. 
However, it is worth noting that the price of API calls is subject to change.\footnote{For example, the price for the davinci GPT-3 engine has been reduced from \$0.06 per 1,000 tokens to \$0.02 per 1,000 tokens. It will probably be further reduced in the future.}

\section{Text Classification Experiments}
\label{sec:setup}

In this section, we present our experimental setup and experimental results for our classification task. 
In \Cref{sec:analysis}, we will demonstrate how to utilize these results for quantitatively analyzing the models' cost/quality trade-offs. 

\subsection{Datasets}
\label{sec:datasets}

\paragraphnew{Existing Datasets.} 
We experiment with the following datasets in the literature: 
AG's News corpus \cite{NIPS2015_250cf8b5}, 20 News Group dataset \cite{Lang95}, and IMDb review (sentiment) dataset \cite{maas-EtAl:2011:ACL-HLT2011}.
These datasets are selected to cover a variety number of target labels, and the scale of these datasets is large enough to cover a 200-shot setting.
The statistics of the above datasets are summarized in \Cref{tb:dataset_stats}. 

\paragraphnew{INCA Information Intent Dataset.}
We also experiment with our internal data, named INCA\_II. 
It is collected from our INCA platform, the goal of which is to collect human responses to surveys for better quantitative market research.
Specifically, given a question $q$ and a response $r$ from the user, we aim at detecting the user's intention in the response.
We assign three labels: ``relevant'' means the user is able to appropriately respond to the given question; ``irrelevant'' means the user is responding with something that is kind of responsive to the given question, however this responsive is not totally relevant to the question; and ``unresponsive'' refers to the situations where the response does not answer the question.

Our dataset contains a collection of 3,000 users' responses from our INCA platform over 101 different questions. 
We use GPT-3 to generate initial labels. These label guesses are then manually checked by one of our own market researchers.

\begin{table}[]
\centering
\resizebox{0.48\textwidth}{!}{
\begin{tabular}{l|cccc|c}
\toprule
Dataset & \# Labels & \# Train & \# Test & Length & Anno. Cost\\\midrule
IMDb & 2 & 25,000 & 25,000 & 233.8 & 0.012 \\
AG News & 4 & 120,000 & 7,600 & 43.9 & 0.0023\\
20 News & 20 & 11,314 & 7,532 & 406.0 & 0.021\\\midrule
INCA\_II & 3 & 2,181 & 819 & 27.9 & 0.072\\
\bottomrule
\end{tabular}
}
\caption{Our dataset statistics. The annotation cost is presented in US dollars per sample.}
\label{tb:dataset_stats}
\end{table}

\subsection{Models}

\paragraphnew{LR and SVM.}
For non-neural models, we experiment with logistic regression (LR) and the support vector machine (SVM).
TF-IDF features extracted from the raw text are used as inputs to the model.\footnote{We build the vocabulary by considering the top 100,000 words ordered by term frequency across the corpus.}

\paragraphnew{Kim CNN.}
We experiment with 1D convolutional neural networks \cite{kim-2014-convolutional}. 
The convolution kernel sizes are 3, 4 and 5-grams with 100 filters for each.  We then minimize cross-entropy loss using Adam \cite{iclr:adam}. The network is trained for 30 epochs with a learning rate of 0.01 and a batch size of 32.

\paragraphnew{BERT.}
We fine-tune a BERT-base model \cite{devlin-etal-2019-bert} using the Adam optimizer with a standard configuration of a learning rate of 2e-5, a batch size of 16, and a maximal sequence length of 256. The model is trained for 10 epochs.

\paragraphnew{T5.}
We also conduct experiments on a T5-base model \cite{JMLR:v21:20-074}.
The model is fine-tuned using the Adam optimizer with a learning rate of 3e-4. The training batch size is set to be 8, the maximum sequence length is 512, and the total training epoch is 10.
To make sure the target labels are not tokenized by the tokenizer, we use the special tokens as targets in our experiments.\footnote{For example, we assign {\tt <extra\_id\_0>} as ``negative'' and {\tt <extra\_id\_1>} as ``positive'' for IMDB dataset.}
For model predictions, we compare the probability for each target label and then choose the largest among those labels as the model outputs.

\paragraphnew{GPT-3.} 
We consider the in-context learning setting when experimenting with GPT-3 model \cite{NEURIPS2020_1457c0d6}.
Pairs of utterances and gold labels as included as in-context examples into the prompt.\footnote{Experiments are done through public APIs of OpenAI GPT-3 with {\tt text-davinci-002} engine in \url{https://openai.com/api/}.} 
The selection of these examples is determined through various iterations in pilot studies.

\begin{table*}[h!]
\centering
\resizebox{0.99\textwidth}{!}{
\begin{tabular}{lrrrrrrrrrrrrrrrrrrrrrr}
\toprule
 \multirow{3}{*}{$K$} & \multicolumn{15}{c}{Total Running Time (s)} & \multicolumn{6}{c}{Accuracy (\%)} \\\cline{2-15}\cline{17-22}
 & \multicolumn{2}{c}{LR} & & \multicolumn{2}{c}{SVM} & & \multicolumn{2}{c}{CNN} & & \multicolumn{2}{c}{BERT} & & \multicolumn{2}{c}{T5} & & \multirow{2}{*}{LR} & \multirow{2}{*}{SVM} & \multirow{2}{*}{CNN} & \multirow{2}{*}{BERT} & \multirow{2}{*}{T5} & \multirow{2}{*}{GPT-3} \\\cline{2-3}\cline{5-6}\cline{8-9}\cline{11-12}\cline{14-15} 
 & Train       & Test  &    & Train       & Test   &    & Train       & Test   &    & Train      & Test    &   & Train       & Test & & & & & & \\\midrule
\multicolumn{22}{c}{\textit{IMDb}} \\
0   & -- & --    &  & -- & --     &  & -- & --     &  & -- & --      &  & --  & --      &  & --   & --   & --   & --   & --   & 85.0 \\
1 & 6.4 & $<$0.1 &   & 6.1 & $<$0.1 &   & 3.7 & 3.5 &   & 58.3 & 406.9 &   & 5810.1 & 595.1 &   & 50.6$_{0.5}$ & 50.9$_{1.8}$ & 51.4$_{1.5}$ & 53.3$_{0.0}$ & 49.7$_{0.0}$ & 95.5 \\
5 & 6.2 & $<$0.1 &   & 6.1 & $<$0.1 &   & 5.5 & 3.6 &   & 60.2 & 409.0 &   & 5704.8 & 533.0 &   & 55.1$_{2.7}$ & 56.4$_{2.3}$ & 54.9$_{2.6}$ & 53.3$_{0.0}$ & 49.7$_{0.0}$ & --\\
10 & 6.3 & $<$0.1 &   & 6.2 & $<$0.1 &   & 4.8 & 3.5 &   & 81.0 & 409.3 &   & 8872.4 & 786.6 &   & 57.5$_{3.3}$ & 58.4$_{0.8}$ & 58.6$_{3.2}$ & 56.4$_{2.8}$ & 49.7$_{0.0}$ & --\\
50 & 6.5 & $<$0.1 &   & 6.3 & $<$0.1 &   & 8.2 & 3.2 &   & 121.8 & 409.5 &   & 4995.9 & 501.0 &   & 68.9$_{1.0}$ & 70.7$_{2.1}$ & 68.8$_{2.3}$ & 80.9$_{5.3}$ & 49.7$_{0.0}$ & --\\
100 & 6.6 & $<$0.1 &   & 6.4 & $<$0.1 &   & 9.9 & 3.2 &   & 156.0 & 407.9 &   & 5595.7 & 482.0 &   & 73.3$_{3.1}$ & 74.5$_{1.2}$ & 77.2$_{1.8}$ & 85.2$_{0.9}$ & 50.0$_{0.1}$ & --\\
200 & 6.7 & $<$0.1 &   & 6.6 & $<$0.1 &   & 14.1 & 3.2 &   & 231.3 & 409.5 &   & 6303.8 & 536.9 &   & 76.8$_{1.9}$ & 77.5$_{1.1}$ & 80.7$_{1.1}$ & 87.6$_{0.5}$ & 90.1$_{0.6}$ & --\\
12,500 & 12.7 & $<$0.1 &   & 10.9 & $<$0.1 &   & 1151.8 & 3.2 &   & 9522.6 & 426.0 &   & 14960.7 & 338.7 &   & 87.9$_{0.0}$ & 87.4$_{0.0}$ & 87.6$_{0.1}$ & 94.0$_{0.2}$ & 94.8$_{0.1}$ & -- \\
\midrule
 \multicolumn{22}{c}{\textit{AG News}} \\
0   & -- & --    &  & -- & --     &  & -- & --     &  & -- & --      &  & --  & --      &  & --   & --   & --   & --   & --   & 76.0 \\
1 & 7.0 & $<$0.1 &   & 6.9 & $<$0.1 &   & 2.3 & 0.5 &   & 49.0 & 51.8 &   & 1541.4 & 143.5 &   & 33.1$_{4.3}$ & 31.6$_{1.1}$ & 43.8$_{7.0}$ & 27.3$_{0.0}$ & 25.1$_{0.0}$ & 86.0\\
5 & 6.8 & $<$0.1 &   & 6.7 & $<$0.1 &   & 3.0 & 0.4 &   & 73.5 & 51.9 &   & 1582.2 & 147.4 &   & 45.8$_{2.6}$ & 45.1$_{2.5}$ & 69.9$_{4.0}$ & 66.2$_{4.0}$ & 25.1$_{0.0}$ & 88.0\\
10 & 7.0 & $<$0.1 &   & 6.7 & $<$0.1 &   & 3.4 & 0.4 &   & 80.7 & 51.9 &   & 1755.3 & 159.1 &   & 53.4$_{2.6}$ & 53.2$_{1.9}$ & 76.2$_{1.4}$ & 81.4$_{2.9}$ & 25.1$_{0.0}$ & 92.0\\
50 & 7.0 & $<$0.1 &   & 6.9 & $<$0.1 &   & 6.6 & 0.5 &   & 100.3 & 51.9 &   & 1891.0 & 160.4 &   & 73.7$_{0.3}$ & 74.1$_{0.6}$ & 84.0$_{0.2}$ & 87.3$_{0.5}$ & 25.8$_{0.4}$ & -- \\
100 & 7.1 & $<$0.1 &   & 6.9 & $<$0.1 &   & 12.8 & 0.5 &   & 140.9 & 51.9 &   & 2030.3 & 160.9 &   & 79.6$_{0.5}$ & 79.5$_{0.8}$ & 85.5$_{0.7}$ & 88.3$_{0.3}$ & 79.6$_{2.0}$ & --\\
200 & 7.3 & $<$0.1 &   & 6.7 & $<$0.1 &   & 20.8 & 0.5 &   & 191.0 & 51.9 &   & 2234.8 & 160.2 &   & 83.0$_{0.5}$ & 83.3$_{0.4}$ & 86.2$_{0.5}$ & 88.9$_{0.4}$ & 89.5$_{0.2}$ & --\\
30,000 & 51.3 & $<$0.1 &   & 13.5 & $<$0.1 &   & 5501.7 & 0.4 &   & 37661.6 & 69.8 &   & 41477.6 & 147.7 &   & 91.7$_{0.0}$ & 92.1$_{0.0}$ & 89.9$_{0.4}$ & 94.6$_{0.1}$ & 94.8$_{0.1}$ & --\\
\midrule
\multicolumn{22}{c}{\textit{20 News Group}} \\
0   & -- & --    &  & -- & --     &  & -- & --     &  & -- & --      &  & --  & --      &  & --   & --   & --   & --   & --   & 30.0 \\
1 & 6.9 & $<$0.1 &   & 5.3 & $<$0.1 &   & 18.9 & 1.6 &   & 756.1 & 141.1 &   & 1240.0 & 124.5 &   & 19.1$_{1.5}$ & 23.4$_{1.1}$ & 10.8$_{1.5}$ & 6.7$_{1.3}$ & 4.2$_{0.0}$ & 40.0 \\
5 & 7.6 & $<$0.1 &   & 5.4 & $<$0.1 &   & 34.6 & 1.8 &   & 888.3 & 138.5 &   & 1718.8 & 161.1 &   & 42.1$_{1.1}$ & 41.1$_{1.6}$ & 34.4$_{4.2}$ & 40.9$_{10.0}$ & 4.2$_{0.0}$ & -- \\
10 & 8.9 & $<$0.1 &   & 5.3 & $<$0.1 &   & 45.1 & 1.9 &   & 916.7 & 137.5 &   & 947.6 & 80.5 &   & 52.6$_{0.8}$ & 52.4$_{1.0}$ & 48.3$_{1.6}$ & 62.7$_{2.0}$ & 4.5$_{0.2}$ & --\\
50 & 12.4 & $<$0.1 &   & 5.8 & $<$0.1 &   & 59.8 & 1.6 &   & 1249.5 & 137.9 &   & 1496.1 & 98.7 &   & 71.1$_{0.6}$ & 72.2$_{0.7}$ & 69.0$_{1.0}$ & 76.5$_{0.9}$ & 62.1$_{1.1}$  & --\\
100 & 13.8 & $<$0.1 &   & 6.2 & $<$0.1 &   & 73.3 & 1.6 &   & 1649.6 & 137.9 &   & 1983.4 & 104.1 &   & 76.1$_{0.3}$ & 77.3$_{0.3}$ & 73.1$_{0.7}$ & 80.1$_{0.7}$ & 77.5$_{0.4}$ & --\\
200 & 21.5 & $<$0.1 &   & 9.7 & $<$0.1 &   & 192.5 & 4.6 &   & 2423.7 & 138.2 &   & 2535.6 & 101.0 &   & 79.5$_{0.2}$ & 81.3$_{0.4}$ & 75.2$_{0.6}$ & 83.4$_{0.3}$ & 82.6$_{0.3}$ & -- \\
Full$^\dagger$ & 48.0 & $<$0.1 &   & 9.7 & $<$0.1 &   & 585.2 & 1.7 &   & 5395.1 & 136.5 &   & 4453.2 & 80.4 &   & 83.3$_{0.0}$ & 85.2$_{0.0}$ & 78.7$_{0.3}$ & 87.4$_{0.1}$ & 86.7$_{0.2}$ & --\\
\midrule
\multicolumn{22}{c}{\textit{INCA Information Intent}} \\
0   & -- & --    &  & -- & --     &  & -- & --     &  & -- & --      &  & --  & --      &  & --   & --   & --   & --   & --   & 46.5 \\
1 & $<$0.1 & $<$0.1 &   & $<$0.1 & $<$0.1 &   & 1.6 & $<$0.1 &   & 57.2 & 11.6 &   & 207.5 & 18.4 &   & 18.7$_{6.7}$ & 38.5$_{15.8}$ & 54.5$_{28.4}$ & 81.1$_{0.0}$ & 87.7$_{0.0}$ & 78.3\\
5 & $<$0.1 & $<$0.1 &   & $<$0.1 & $<$0.1 &   & 1.5 & $<$0.1 &   & 59.4 & 12.5 &   & 210.8 & 18.0 &   & 27.1$_{15.2}$ & 39.6$_{9.0}$ & 60.1$_{14.8}$ & 81.1$_{0.0}$ & 87.7$_{0.0}$ & 87.1 \\
10 & 0.1 & $<$0.1 &   & $<$0.1 & $<$0.1 &   & 1.7 & $<$0.1 &   & 70.1 & 11.2 &   & 216.9 & 17.2 &   & 21.2$_{6.8}$ & 30.5$_{7.2}$ & 62.5$_{17.2}$ & 62.7$_{12.7}$ & 87.7$_{0.0}$ & -- \\
20 & 0.1 & $<$0.1 &   & $<$0.1 & $<$0.1 &   & 2.4 & 0.1 &   & 95.4 & 11.1 &   & 241.4 & 17.4 &   & 33.2$_{8.8}$ & 28.8$_{8.7}$ & 75.8$_{17.4}$ & 62.2$_{13.0}$ & 87.7$_{0.0}$ & -- \\
50 & 0.1 & $<$0.1 &   & $<$0.1 & $<$0.1 &   & 6.4 & 0.1 &   & 96.6 & 8.3 &   & 311.7 & 18.1 &   & 25.6$_{3.1}$ & 32.8$_{8.3}$ & 68.5$_{4.5}$ & 74.8$_{5.3}$ & 87.7$_{0.0}$ & --\\
Full$^\dagger$ & 0.3 & $<$0.1 &   & 0.1 & $<$0.1 &   & 66.8 & $<$0.1 &   & 925.8 & 11.1 &   & 1632.1 & 18.3 &   & 87.7$_{0.0}$ & 87.8$_{0.0}$ & 86.9$_{1.1}$ & 88.6$_{0.3}$ & 90.2$_{0.8}$ & -- \\
\bottomrule
\end{tabular}}
\caption{Running time and model accuracy in $K$-shot learning for different datasets. 
Due to budget issues, we only run GPT-3 on a subset of 200 examples from test set of IMDb, AG News, and 20 News Group datasets. 
All reported accuracy is averaged over 5 runs with subscripts denoting the standard deviation. 
$^\dagger$As the maximum number of samples is different for each category, we use ``Full'' to denote the model is trained on the whole training set.}
\label{tb:main_results}
\end{table*}

\subsection{Experimental Setup}

In order to understand the model classification accuracy along with its costs under different scenarios, we consider a few-shot setting in this work: for $N$ classes, we randomly sample a number of $K$ examples from each class.
It thus simulates the scenario when there is a fixed budget, for example, the number of available annotations.
In our experiments, we consider $K = \{1, 5, 10, 50, 100, 200\}$.\footnote{For INCA\_II dataset, $K=\{1, 5, 10, 20, 50\}$ and the full train set are considered due to dataset size.} 
For reference and comparison purposes, we also include the experimental results for models trained on the full training set. 

We use accuracy as the main metric for evaluating the performances of the classifiers.
For IMDb, AG News, and 20 News group, we use the established splits for the train set and test set.
A subset from the train set is used as the validation set.
For our INCA Information Intent dataset, we randomly divide the total 101 questions into 65 train questions, 5 validation questions, and the 31 left as test questions. We then collect the users' responses to these questions to generate the train/dev/test splits.
We select the hyperparameters on the validation set.
All our experiments are done in an in-house NVIDIA RTX A6000 GPU with 48GB memory.

\subsection{Classification Accuracy Results}
\label{subsec:classifier_performance}

We experiment with different choices of shots $K$ and present our experimental results on the test set in \Cref{tb:main_results}. 
For each configuration, we report the averages across five random trials.

We observe that BERT and GPT-3 generally do not suffer from the cold start problem, i.e., they work reasonably well in zero or few shots; while the T5-base model normally requires a certain number of training examples before showing decent accuracy.
We also do not observe a significant difference for BERT-base and T5-base when the model is trained on the full training set.

Specific to our INCA Information Intent dataset, due to its label imbalance nature,\footnote{In our platform, most users' responses are valid and thus marked as ``relevant''.} a simple majority baseline predicting all test samples as ``relevant'' yields an 87.7\% accuracy in our test set.
We observe that only BERT and T5 are able to outperform this baseline while other models are not.

\section{Evaluating Cost/Quality Trade-offs}
\label{sec:analysis}

\begin{figure*}[h!]
\centering
\begin{minipage}[h]{0.47\linewidth}
\centering
\includegraphics[width=0.99\linewidth]{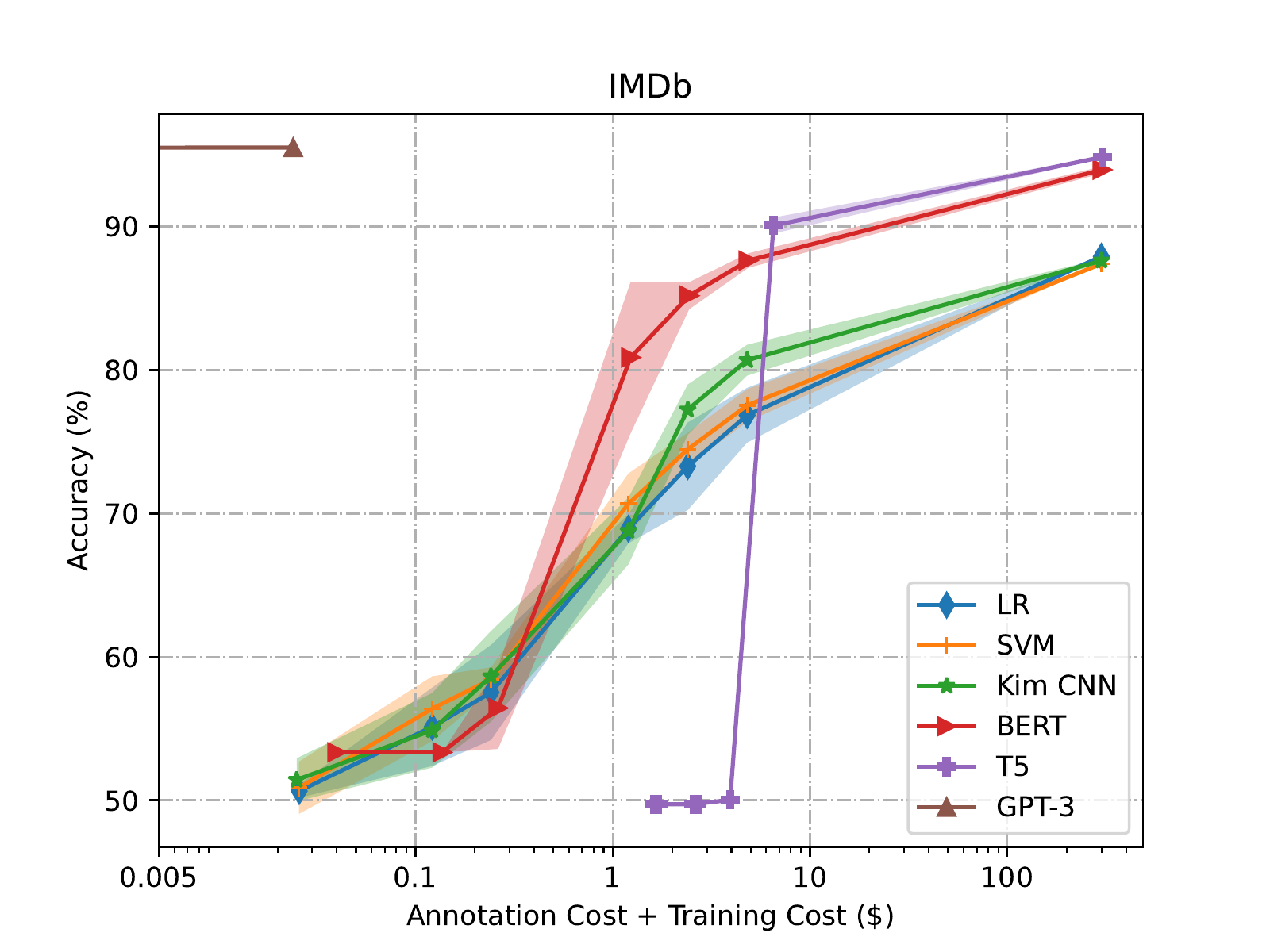}
\end{minipage}
\begin{minipage}[h]{0.47\linewidth}
\centering
\includegraphics[width=0.99\linewidth]{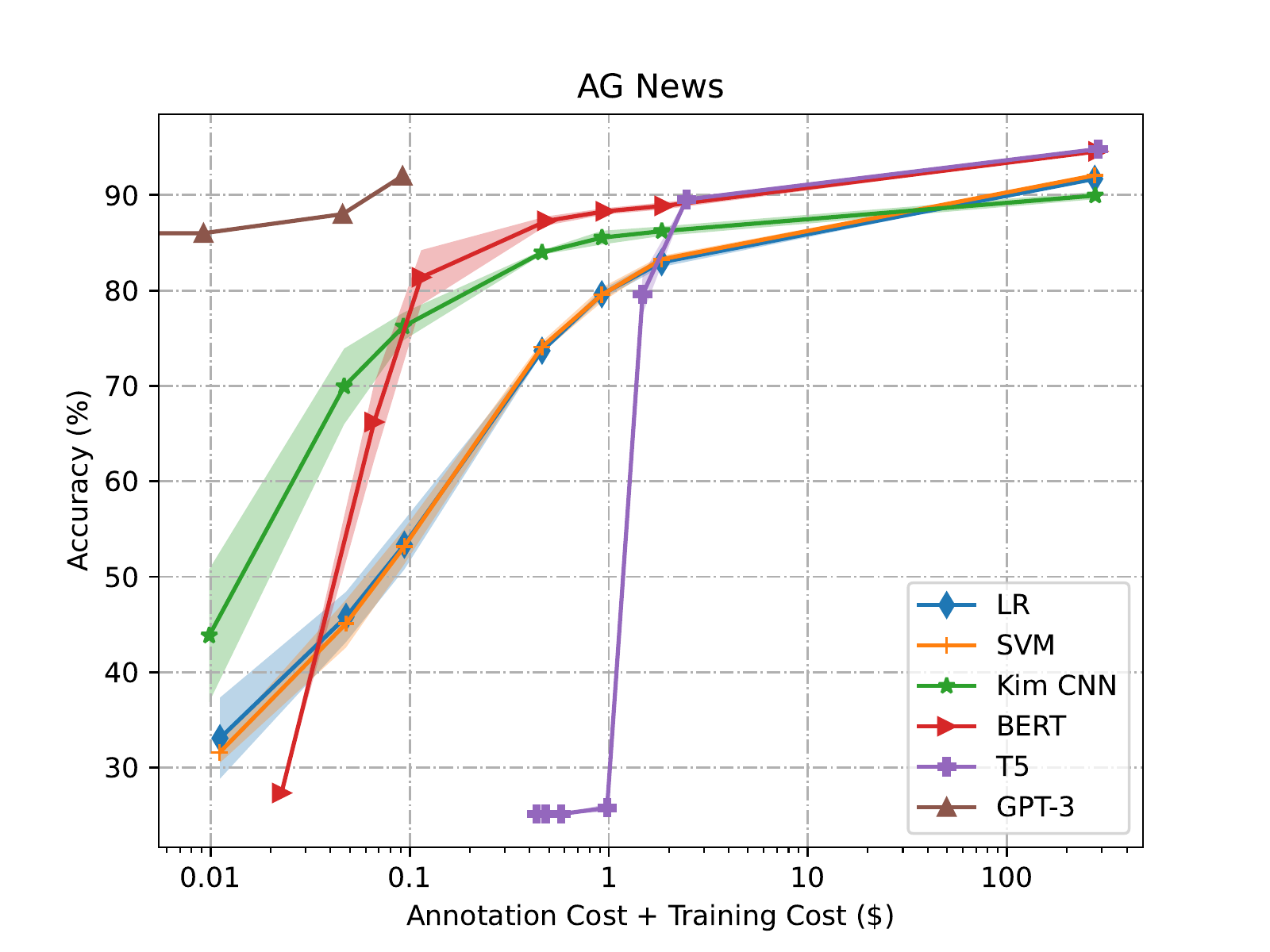}
\end{minipage}
\\
\begin{minipage}[h]{0.47\linewidth}
\centering
\includegraphics[width=0.99\linewidth]{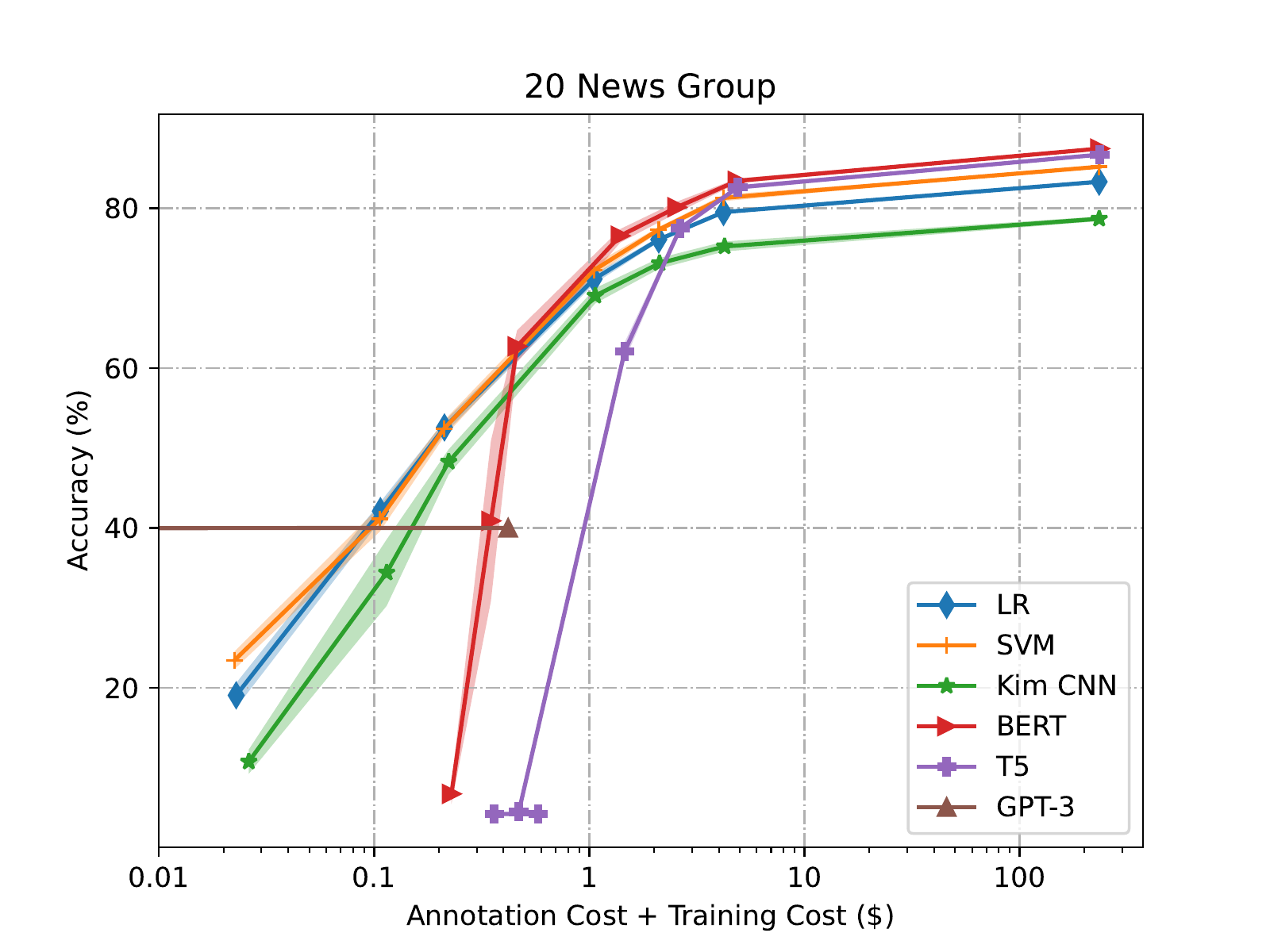}
\end{minipage}
\begin{minipage}[h]{0.47\linewidth}
\centering
\includegraphics[width=0.99\linewidth]{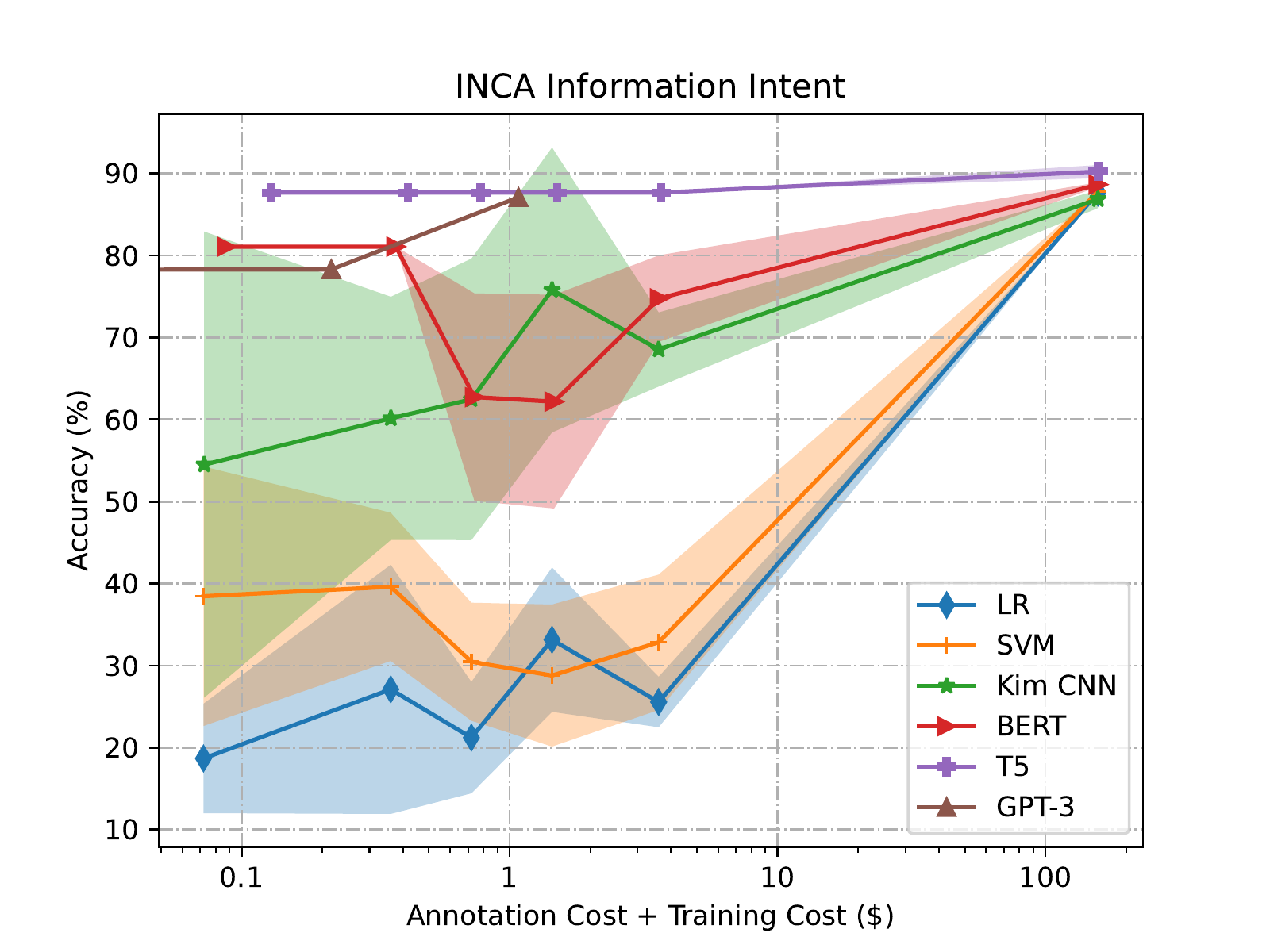}
\end{minipage}
\caption{Models' classification accuracy (\%) under different annotation and training costs (US Dollars).}
\label{fig:cost_quality_tradeoff}
\end{figure*}

\subsection{Estimating Cost}
\label{sec:cost_estimate}

\paragraphnew{Estimating Annotation Cost.}
We first estimate the annotation cost to obtain labels for our used datasets.
We use the suggested price of a human labeler from Amazon Mechanical Turk, which is roughly \$0.012 per around 200 words for the classification task.\footnote{\url{https://aws.amazon.com/sagemaker/data-labeling/pricing/}}
The annotation cost for IMDb, AG News, and 20 News Group datasets is then estimated using the average number of tokens multiplied by the above price in \Cref{tb:dataset_stats}.\footnote{The original labels for these datasets are automatically generated. In this work, we only consider one human labeler for these three datasets. The annotation price for other datasets might be higher, e.g., using multiple labelers.}

For our INCA Information Intent dataset, the annotation cost consists of the initial cost using GPT-3 and the manual efforts from our in-house domain expert (details discussed in \Cref{sec:datasets}).
The total time cost is 5.5 hours by the researcher and the total annotation cost is thus estimated to be \$0.072 per example.\footnote{The hourly pay rate is estimated based on the per hour average compensation of a market researcher provided by Statistics Canada at: \url{https://www150.statcan.gc.ca/t1/tbl1/en/tv.action?pid=3610048906}.}
Compared to having the in-house domain expert annotate from scratch, we find this method seems to reduce the total cost for the annotation.

\paragraphnew{Estimating Model Cost.}
We make the approximated cost for the train/fine-tune cost and inference cost based on the time length reported in \Cref{tb:main_results}.
To ensure a fair and accurate estimate, we use the price for renting the same type of GPU from Jarvislabs.ai for our cost calculation, which is \$0.99 per hour.\footnote{\url{https://jarvislabs.ai/pricing/}}
The price for a GPT-3 API call with {\tt text-davinci-002} engine is based on the combined input and output length, with a price of \$0.02 per 1000 tokens. The cost calculation for GPT-3 is thus exact.

There is a pre-training cost for BERT-base, T5-base, and GPT-3 models. 
Here as we do not pre-train our in-domain models, we do not include the pre-training cost as a part of the model cost. 

\subsection{Case Studies}

With all the estimates and measurements above, we are now ready to answer the following question: suppose we need to make a certain number of inferences, how can we select the appropriate model to not only make sure our anticipated accuracy can be met, but also to minimize the total cost?

To this end, we visualize the different models' performance changes over the total costs calculated in \Cref{sec:cost_estimate}. 
We then calculate the estimated cost under different expected accuracy given a total number of 100,000 samples for inference, based on the cost/quality trade-off curves in \Cref{fig:cost_quality_tradeoff}.
This sample size is roughly the monthly rate needed for intent classification in our INCA platform,\footnote{Based on our internal statistics, there are 3,000 $\sim$ 5,000 requests per day, which makes 100,000 per month.} reflecting the actual amount of requests one would encounter in real practice.

Our results are presented in \Cref{tb:100k_breakdowns}. 
We observe that although non-neural models such as logistic regression and support vector machine have nearly zero inference cost, they require a large amount of data points to reach a decent accuracy. Therefore non-neural models can be cost-effective for large-scale inference, the need for increased annotation cost can result in higher total costs for a more moderate inference volume.
As demonstrated in \Cref{subsec:classifier_performance}, GPT-3 usually does not require a huge amount of annotated data and performs well in few-shot settings. It can reduce the annotation and training costs, but will induce a large cost when having a large-scale test set for inference.
Moreover, few-shot GPT-3 may underperform in classification accuracy for certain tasks.

We further select IMDb and our own INCA Information Intent dataset and calculate the estimated total cost under a different number of test samples and expected accuracy in \Cref{tb:case_study}.
We observe that for the IMDb dataset, if a high accuracy level (90\%) is required but there is only a small sample size ($<$500), then GPT-3 would be the most cost-effective; while a high accuracy level (90\%) and a larger sample size ($>$500) would lead to the T5 model.
BERT seems to be the most cost-effective if we want to trade classification accuracy (a relatively low accuracy such as 85\% or lower is acceptable) for a large number of samples for inference.

\begin{table}[]
\centering
\resizebox{0.48\textwidth}{!}{
\begin{tabular}{lrrlrrlrr}
\toprule
 & \multicolumn{2}{c}{83\%} &  & \multicolumn{2}{c}{85\%} &  & \multicolumn{2}{c}{88\%} \\\cline{2-3} \cline{5-6} \cline{8-9}
 & \multicolumn{1}{c}{Anno.+Tr.} & \multicolumn{1}{c}{Infer.} &  & \multicolumn{1}{c}{Anno.+Tr.} & \multicolumn{1}{c}{Infer.} &  & \multicolumn{1}{c}{Anno.+Tr.} & \multicolumn{1}{c}{Infer.} \\\midrule
\multicolumn{9}{c}{\textit{IMDb}} \\
LR & 169.16 & $<$0.01 &  & 222.57 & $<$0.01 &  & -- & -- \\
SVM & 168.49 & $<$0.01 &  & 228.32 & $<$0.01 &  & -- & -- \\
CNN & 103.64 & $<$0.01 &  & 188.87 & $<$0.01 &  & -- & -- \\
BERT & 1.83 & 0.45 &  & 2.39 & 0.45 &  & 22.66 & 0.45 \\
T5 & 6.08 & 0.59 &  & 6.21 & 0.59 &  & 6.40 & 0.59 \\
GPT-3 & $<$0.01 & 661.88 &  & $<$0.01 & 661.88 &  & 0.01 & 1937.76 \\
\midrule
\multicolumn{9}{c}{\textit{AG News}} \\
LR & 3.08 & $<$0.01 &  & 65.84 & $<$0.01 &  & 159.99 & $<$0.01 \\
SVM & 1.78 & $<$0.01 &  & 55.87 & $<$0.01 &  & 149.34 & $<$0.01 \\
CNN & 0.42 & $<$0.01 &  & 0.76 & $<$0.01 &  & 133.75 & $<$0.01 \\
BERT & 0.22 & 0.20 &  & 0.34 & 0.20 &  & 0.82 & 0.20 \\
T5 & 1.81 & 0.53 &  & 2.01 & 0.53 &  & 2.30 & 0.53 \\
GPT-3 & 0.01 & 646.08 &  & 0.01 & 646.08 &  & 0.05 & 1645.08 \\
\midrule
\multicolumn{9}{c}{\textit{20 News Group}} \\
LR & 215.82 & $<$0.01 &  & -- & -- &  & -- & -- \\
SVM & 105.84 & $<$0.01 &  & 223.79 & $<$0.01 &  & -- & -- \\
CNN & -- & -- &  & -- & -- &  & -- & -- \\
BERT & 4.58 & 0.50 &  & 95.83 & 0.50 &  & -- & -- \\
T5 & 28.09 & 0.39 &  & 140.37 & 0.39 &  & -- & -- \\
GPT-3 & -- & -- &  & -- & -- &  & -- & -- \\
\midrule
\multicolumn{9}{c}{\textit{INCA Information Intent}} \\
LR & / & / &  & / & / &  & -- & -- \\
SVM & / & / &  & / & / &  & -- & -- \\
CNN & / & / &  & / & / &  & -- & -- \\
BERT & / & / &  & / & / &  & 150.14 & 0.37 \\
T5 & / & / &  & / & / &  & 23.61 & 0.60 \\
GPT-3 & / & / &  & / & / &   & -- & -- \\
\bottomrule
\end{tabular}
}
\caption{Cost breakdowns (annotation cost+train cost versus inference cost; in US dollars) given 100,000 samples for inference. For INCA Information Intent dataset, we mark the anticipated accuracy below 87.7\% as ``/'', as a majority baseline will have such accuracy (discussed in \Cref{subsec:classifier_performance}). Blocks marked as ``--'' are those accuracy levels that a certain model can not reach.}
\label{tb:100k_breakdowns}
\end{table}

\begin{table}[h!]
\centering
\resizebox{0.48\textwidth}{!}{
\begin{tabular}{lrrrrrrr}
\toprule
 Sample Size & \multicolumn{1}{r}{200} & \multicolumn{1}{r}{500} & \multicolumn{1}{r}{1,000} & \multicolumn{1}{r}{5,000} & \multicolumn{1}{r}{10,000} & \multicolumn{1}{r}{50,000} & \multicolumn{1}{r}{100,000} \\\midrule
\multicolumn{8}{c}{\textit{IMDb}} \\
\multicolumn{8}{l}{\underline{\textbf{75\% Acc.}}}\\
LR & 3.56	& 3.56	& 3.56	& 3.56	& 3.56	& 3.56	& 3.56 \\
SVM & 2.82	& 2.82	& 2.82	& 2.82	& 2.82	& 2.82	& 2.82 \\
CNN & 2.09	& 2.09	& 2.09	& 2.09	& 2.09	& 2.09	& 2.09 \\
BERT & 1.00	& 1.00	& 1.00	& 1.02	& 1.05	& 1.23	& 1.45 \\
T5 & 5.56	& 5.56	& 5.56	& 5.59	& 5.62	& 5.85	& 6.15 \\
GPT-3 & 1.32	& 3.31	& 6.62	& 33.09	& 66.19	& 330.94	& 661.88 \\
\multicolumn{8}{l}{\underline{\textbf{85\% Acc.}}}\\
LR & 222.57 & 222.57 & 222.57 & 222.57 & 222.57 & 222.57 & 222.57 \\
SVM & 228.32 & 228.32 & 228.32 & 228.32 & 228.32 & 228.32 & 228.32 \\
CNN & 188.87 & 188.87 & 188.87 & 188.87 & 188.87 & 188.87 & 188.87 \\
BERT & 2.39 & 2.40 & 2.40 & 2.42 & 2.44 & 2.62 & 2.85 \\
T5 & 6.21 & 6.21 & 6.21 & 6.23 & 6.26 & 6.50 & 6.80 \\
GPT-3 & 1.32 & 3.31 & 6.62 & 33.09 & 66.19 & 330.94 & 661.88\\
\multicolumn{8}{l}{\underline{\textbf{90\% Acc.}}}\\
LR & -- & -- & -- & -- & -- & -- & -- \\
SVM & -- & -- & -- & -- & -- & -- & -- \\
CNN & -- & -- & -- & -- & -- & -- & -- \\
BERT & 116.75 & 116.75 & 116.75 & 116.77 & 116.79 & 116.97 & 117.20 \\
T5 & 6.53 & 6.53 & 6.54 & 6.56 & 6.59 & 6.83 & 7.12 \\
GPT-3 & 3.89 & 9.70 & 19.39 & 96.90 & 193.79 & 968.89 & 1,937.77 \\
\multicolumn{8}{l}{\underline{\textbf{93\% Acc.}}}\\
LR & -- & -- & -- & -- & -- & -- & -- \\
SVM & -- & -- & -- & -- & -- & -- & -- \\
CNN & -- & -- & -- & -- & -- & -- & -- \\
BERT & 257.87 & 257.87 & 257.88 & 257.90 & 257.92 & 258.10 & 258.32 \\
T5 & 189.37 & 189.37 & 189.37 & 189.40 & 189.43 & 189.67 & 189.96 \\
GPT-3 & 3.89 & 9.71 & 19.40 & 96.91 & 193.79 & 968.90 & 1,937.78
\\\midrule
\multicolumn{8}{c}{\textit{INCA Information Intent}} \\
\multicolumn{8}{l}{\underline{\textbf{88\% Acc.}}}\\
LR & -- & -- & -- & -- & -- & -- & -- \\
SVM & -- & -- & -- & -- & -- & -- & -- \\
CNN & -- & -- & -- & -- & -- & -- & -- \\
BERT & 150.15 & 150.15 & 150.15 & 150.16 & 150.18 & 150.33 & 150.51 \\
T5 & 23.61 & 23.61 & 23.61 & 23.64 & 23.67 & 23.91 & 24.21 \\
GPT-3 & -- & -- & -- & -- & -- & -- & -- \\
\multicolumn{8}{l}{\underline{\textbf{90\% Acc.}}}\\
LR & -- & -- & -- & -- & -- & -- & -- \\
SVM & -- & -- & -- & -- & -- & -- & -- \\
CNN & -- & -- & -- & -- & -- & -- & -- \\
BERT & -- & -- & -- & -- & -- & -- & -- \\
T5 & 143.57 & 143.57 & 143.57 & 143.60 & 143.63 & 143.87 & 144.17 \\
GPT-3 & -- & -- & -- & -- & -- & -- & -- \\
\bottomrule
\end{tabular}
}
\caption{Total cost (in US dollars) needed for different models under a fixed number of points for prediction and an anticipated accuracy.}
\label{tb:case_study}
\end{table}

\section{Related Work}

Text classification is a classical problem in natural language processing and has been extensively studied.
We refer readers to recent surveys such as \citet{10.1145/3495162} and \citet{10.1145/3439726} for a complete review of this task. 
In this work, we experiment with three benchmark datasets and our experimental results are comparable to those reported in the literature. We further discuss the cost/quality trade-offs beyond those established results.

Regarding the estimation for the model cost, it is now encouraged to report the total computational budgets (e.g., GPU hours) and payment information for the human annotators.\footnote{For example, ACL rolling review suggests a checklist that covers these aspects: \url{https://aclrollingreview.org/responsibleNLPresearch/}. }
It is thus possible to calculate the cost based on the method outlined in \Cref{sec:factors}.
For example, \citet{10.1145/3381831} list the training time that is publicly released for BERT-large, XLNet, AlphaGo and calculate the corresponding cost of these models.
\citet{DBLP:journals/corr/abs-2004-08900} also review the training cost of large-scale language models. They discuss the drivers of these costs and provide a forecast for the future trend of the costs. We also notice some prior works on analyzing the environmental impact of language models using carbon footprint \cite{10.1145/3381831}.
However, it is beyond the scope of this paper.

Our work can be treated as trying to train models under fixed constraints.
To the best of our knowledge, there is limited work that studies this topic.
\citet{izsak-etal-2021-train} analyze how to train a BERT model with an academic budget using a single low-end deep learning server in 24 hours.
\citet{bai-etal-2021-pre} conduct analyses towards domain adaption under a fixed budget.

\section{Conclusion}

In this work, we present an empirical study of the cost/quality trade-offs for the text classification task.
We experiment with a number of representative classification models, including logistic regression, support vector machine, convolutional neural network, BERT, T5, and GPT-3.
We then provide an analysis of accessing the estimated total cost (including annotation cost, training cost, and inference cost), given a fixed number of data points for prediction and an anticipated classifier accuracy.
Our results suggest that the actual choices of the model are subject to various factors and industry practitioners need to determine the factors of their top priority when making such decisions.

\bibliography{custom}
\bibliographystyle{acl_natbib}

\end{document}